\documentclass[letterpaper, 10 pt, conference]{ieeeconf}  

\IEEEoverridecommandlockouts                              

\usepackage[table]{xcolor}
\usepackage{graphicx}
\usepackage{caption}
\usepackage{amsmath}
\usepackage{amssymb}
\usepackage{amsfonts}
\usepackage{booktabs}
\usepackage{multirow}
\usepackage{tabularx}
\usepackage{xspace}
\usepackage{url}
\usepackage{bmpsize}
\usepackage{subcaption}
\usepackage{bm}
\usepackage{comment}
\usepackage{listings}
\usepackage{tcolorbox}
\usepackage{algorithm}
\usepackage{algorithmicx}
\usepackage{algcompatible}
\usepackage{algpseudocode}
\usepackage{adjustbox}
\usepackage[whole]{bxcjkjatype} 
\let\labelindent\relax
\usepackage{enumitem}

\newcommand{\fref}[1]{Fig.~\ref{#1}}

\newcommand{\sref}[1]{Sec.~\ref{#1}}
\newcommand{\Sref}[1]{Section~\ref{1}}

\newcommand{\etal}{\emph{et al.}\xspace}

\newcommand{\baselineomni}{\emph{OmniGlue}}

\graphicspath{{./figures_general_insertion/}}

\begin{document}

    \title{\LARGE \bf
    Touch2Insert: Zero-Shot Peg Insertion\\
    by Touching Intersections of Peg and Hole}
    \author{Masaru Yajima$^{1}$, Yuma Shin$^{1}$, Rei Kawakami$^{1}$, Asako Kanezaki$^{1}$, Kei Ota$^{2}$
    \thanks{$^{1}$Masaru Yajima, Yuma Shin, Rei Kawakami, and Asako Kanezaki are with Institute of Science Tokyo, Tokyo, Japan.}
    \thanks{$^{2}$Kei Ota is with Mitsubishi Electric, Kanagawa, Japan. {\tt\small Ota.Kei@ds.MitsubishiElectric.co.jp}}
    }

    \maketitle
    \begin{abstract}
    Reliable insertion of industrial connectors remains a central challenge in robotics, requiring sub-millimeter precision under uncertainty and often without full visual access. Vision-based approaches struggle with occlusion and limited generalization, while learning-based policies frequently fail to transfer to unseen geometries. To address these limitations, we leverage tactile sensing, which captures local surface geometry at the point of contact and thus provides reliable information even under occlusion and across novel connector shapes. Building on this capability, we present \emph{Touch2Insert}, a tactile-based framework for arbitrary peg insertion. Our method reconstructs cross-sectional geometry from high-resolution tactile images and estimates the relative pose of the hole with respect to the peg in a zero-shot manner. By aligning reconstructed shapes through registration, the framework enables insertion from a single contact without task-specific training. To evaluate its performance, we conducted experiments with three diverse connectors in both simulation and real-robot settings. The results indicate that Touch2Insert achieved sub-millimeter pose estimation accuracy for all connectors in simulation, and attained an average success rate of 86.7\% on the real robot, thereby confirming the robustness and generalizability of tactile sensing for real-world robotic connector insertion.
    
\end{abstract}

    \section{Introduction}\label{sec:introduction}
Tactile information is an essential modality that enables humans to accurately perceive object shape and pose, providing the basis for dexterous manipulation \cite{Li2020TactileReview}.

Tactile sensing directly captures local geometry at the contact surface, inherently avoiding occlusion and enabling high-resolution measurements.
Such properties
are particularly important for recognizing objects with fine or intricate geometries, such as the cross-sections of industrial connectors \cite{tegin2005tactile}.
A familiar example is identifying a USB-C port on the back of a monitor purely by touch and inserting the cable without visual guidance.

While such insertion tasks are relatively easy for humans, they are far from trivial for robots. The central challenge lies in accurately estimating the hole's geometry and relative pose, which is a prerequisite for successful insertion.

Insertion tasks for industrial connectors require extremely small tolerances, and achieving success demands pose estimation at sub-millimeter precision. Despite recent advances in vision and control, robotic systems still struggle to meet these requirements, especially when dealing with connectors that exhibit complex or irregular cross-sectional geometries.

\begin{figure}[t]
    \centering
    \includegraphics[width=\linewidth]{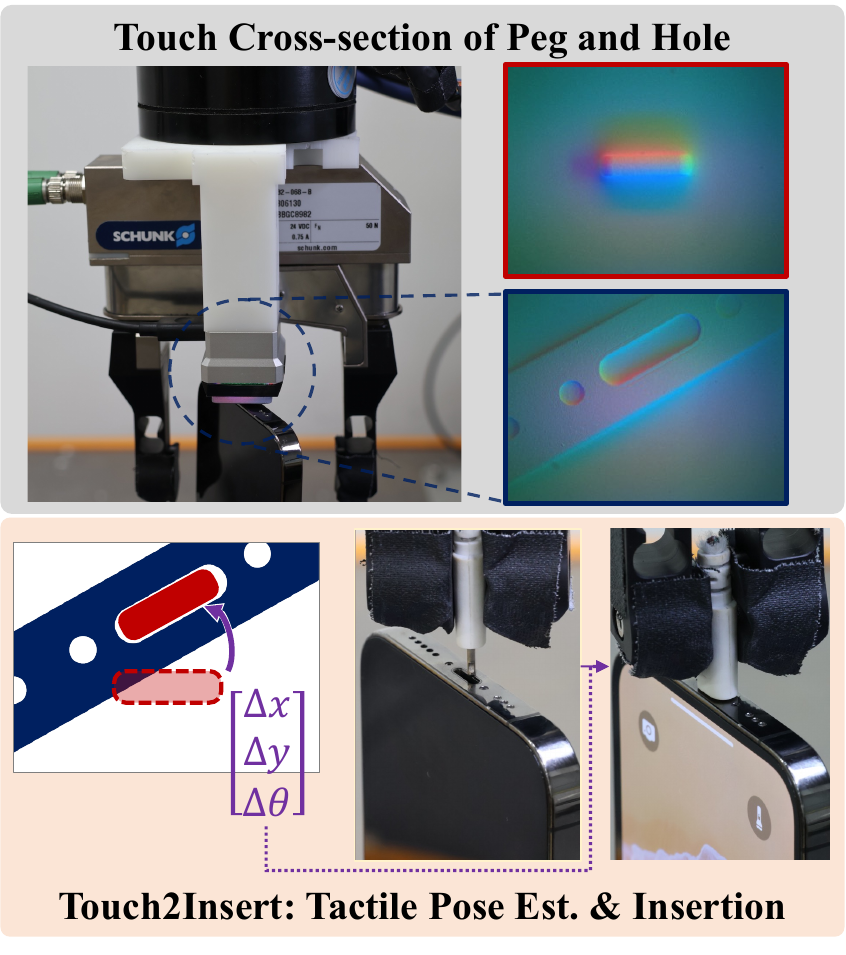}
    \caption{
    This paper addresses the problem of inserting an \emph{arbitrary} peg into an \emph{unknown} hole without any prior knowledge of their shapes or types. We propose a novel framework in which the robot makes contact with the cross-sections of the peg and hole, and estimates the hole's relative pose with respect to the peg in $\mathrm{SE}(2)$ in a zero-shot manner from 3D point clouds obtained from tactile images.
    }
    \vspace{-3mm}
    \label{fig:intro_fig}
\end{figure}

To better frame these challenges, insertion tasks are commonly decomposed into two phases~\cite{xu2019compare,jiang2022areview}:
(1) a \textit{search} phase for identifying the position and orientation of the hole, and
(2) an \textit{insertion} phase in which the robot performs the actual insertion through control.
In this study, we particularly focus on the former exploration phase, since accurate localization of the hole enables reliable insertion using only stiffness control and removes the need for auxiliary exploratory motions such as spiral search~\cite{chhatpar2001search}.

Several studies have investigated insertion using tactile information~\cite{dong2021tactile,kim2022active,dong2019tactile}.
However, these methods generally rely on task-specific training and remain effective only for simple geometries such as rectangular or cylindrical shapes.
In addition, because they use tactile images purely as feedback and compensate for pose errors through repeated interactions, the insertion process becomes slow, inefficient, and unsuitable for connectors with complex cross-sections.
These limitations underscore the need for a fundamentally different approach that directly extracts geometric information from tactile sensing.

Our key idea is to infer the cross-sectional shapes of both the peg and the hole from tactile images and directly estimate their relative pose in $\mathrm{SE}(2)$ space through point cloud registration.
Unlike prior approaches that iteratively compensate for errors, our framework achieves zero-shot pose estimation from a single contact and scales to complex, unseen connector geometries without task-specific training.

We validate our approach through both simulation and real-world experiments.
In simulation, we quantitatively confirm that the hole pose can be estimated with high accuracy across multiple connector types, achieving sub-millimeter precision.
In real-robot experiments, our method achieves an 86.7\% insertion success rate over three different industrial connectors, demonstrating its high performance even under the stringent tolerance requirements of industrial connectors.

Our contributions are summarized as follows:

\begin{enumerate}
\item We propose \emph{Touch2Insert}, a tactile-based peg insertion framework that treats tactile images as geometric observations, reconstructs cross-sectional shapes of the peg and hole, and estimates their relative pose in $\mathrm{SE}(2)$. This framework requires no prior knowledge, achieves zero-shot generalization, and scales to complex connector geometries.
\item We validate the proposed method through both simulation and real-robot experiments, demonstrating accurate connector insertion under stringent industrial tolerances.
\end{enumerate}

    \section{Related Work}\label{sec:related_work}
\textbf{Pose Estimation with Vision Sensors.}
Accurate pose estimation of the target hole is critical for successful peg insertion, as it enables either direct insertion or a significantly reduced search space. Existing approaches often depend on predefined object categories~\cite{gao2021kpam2}, shape priors derived from 3D CAD models~\cite{litvak2019learning}, or extensive task-specific training~\cite{zhang2023vision}, which restrict their applicability to unseen objects. Recent methods leveraging Vision–Language Models~\cite{yajima2025zeroshot} improve generalization, but purely vision-based techniques remain inherently vulnerable to lighting variations and occlusions.
Our approach instead achieves accurate and generalizable pose estimation without requiring object meshes, category definitions, or costly task-specific training. By directly capturing surface geometries through tactile sensing, it avoids these fundamental limitations of vision-based methods and provides a more robust and scalable solution for real-world peg insertion.

\textbf{Pose Estimation with Tactile Sensors.}
High–resolution tactile images enable precise pose inference, and prior work has mainly used them for nearest-neighbor matching against precomputed images or point clouds, requiring CAD models and heavy offline computation. Tac2Pose~\cite{bauze2021tactile, bauza2022tac2pose} matches depth images of gel deformations to rendered candidates, while Yang \etal~\cite{yang2024in-hand} use costly point cloud registration, and MidasTouch~\cite{suresh2022midastouch} aggregates observations with a particle filter.
In contrast, our method directly estimates the relative pose between pegs and holes from tactile contact, eliminating reliance on CAD models, pre-rendering, or pre-collected datasets. This enables robust performance on novel objects and scalability in real-world settings. The closest prior work, Tactile-Filter~\cite{ota2023tactile}, avoids CAD but still requires object-specific data collection, whereas our approach generalizes without such preparation.

    \section{Problem Statement}

We address the problem of \emph{arbitrary peg insertion}, where a robot must autonomously estimate the relative pose of the hole with respect to the peg with sub-millimeter accuracy to achieve reliable insertion. Such precision is essential for industrial connectors, whose tight tolerances make insertion failure highly likely without accurate alignment. While vision can provide a coarse estimate of the hole location, its accuracy is insufficient for this task, and the challenge becomes even greater when the connector is partially or fully occluded by obstacles, rendering purely vision-based methods ineffective.

In this work, we formulate peg insertion as the problem of estimating the relative pose of the hole with respect to the peg in $\mathrm{SE}(2)$, using a vision-based tactile sensor. Upon contact, the sensor captures cross-sectional information of both peg and hole, which is then used to guide the insertion.

The problem is considered under the following assumptions:
\begin{enumerate}
    \item The pose of the peg at the time of grasping is known.
    \item The entire hole lies within the sensing range of the tactile sensor. 
    \item The approximate location of the hole is known.
\end{enumerate}

The first assumption corresponds to the peg being fixed by a jig, so that its pose at the time of grasping is precisely known.
The second assumption ensures that, upon contact, the tactile sensor can observe all or most of the hole’s cross-section.
The third assumption is satisfied by an external vision system that provides the approximate hole location.
In this study, we specifically focus on accurate hole pose estimation, while the subsequent insertion step is performed using existing controllers without requiring additional exploratory motions.

\begin{figure}[t]
    \centering
    \includegraphics[width=\linewidth]{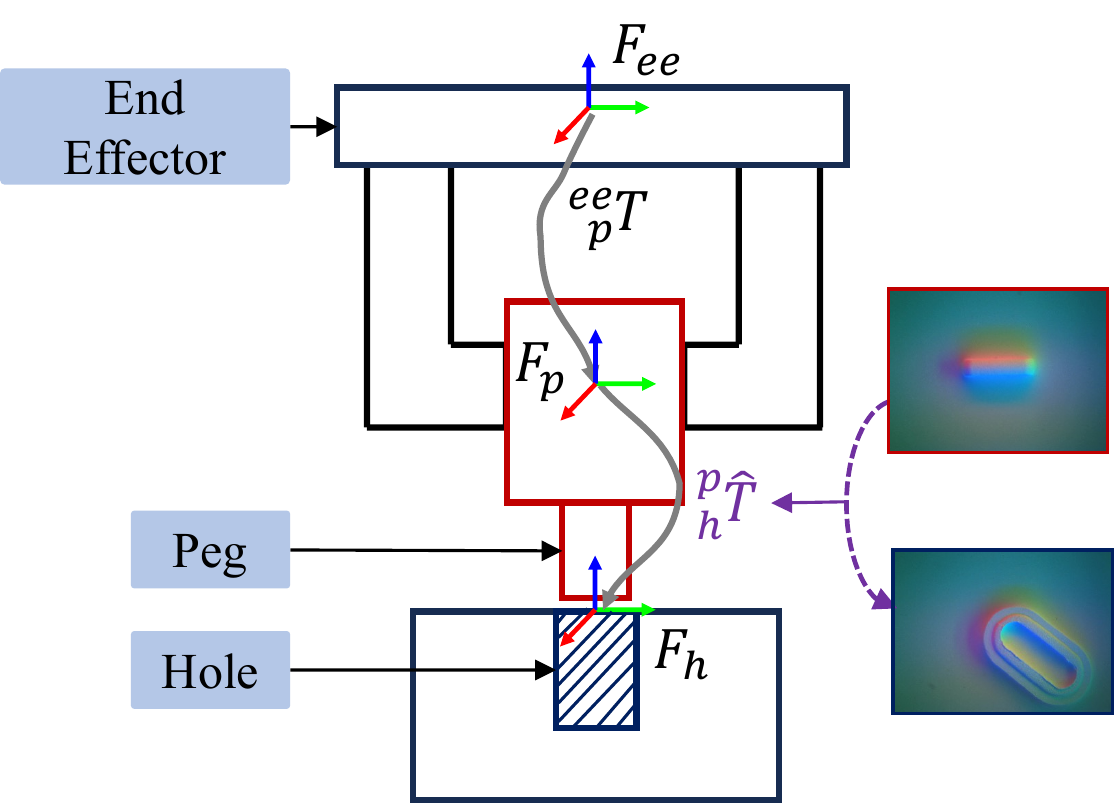}
    \caption{
        \textbf{Definition of coordinates.}
        The coordinate frames are defined as follows: $F_{\mathrm{e}e}$ denotes the end-effector frame, $F_{\mathrm{ee}}$ the frame of the grasped peg, and $F_{\mathrm{h}}$ the hole frame.
        The objective is to estimate the relative pose of the hole with respect to the peg, $^{\mathrm{p}}_{\mathrm{h}}\hat{T}$, which inevitably contains estimation noise.
        This transformation is then incorporated into the coordinate conversion described in Eq.~\ref{transformation}, yielding the estimated end-effector pose in the world frame. Finally, this pose is issued to the robot as a command for execution.
    }
    \label{fig:coordinates}
\end{figure}

The objective of this study is to estimate the $\mathrm{SE}(2)$ pose of the endeffector for insertion in the world coordinate, denoted as $^\mathrm{w}_\mathrm{ee}\hat{T}$, and to perform the insertion based on this estimate.
To this end, we first estimate the relative pose of the hole with respect to the peg, $^\mathrm{p}_\mathrm{h}\hat{T}$, and then transform it into the pose of the end-effector in the world coordinate, $^\mathrm{w}_\mathrm{ee}\hat{T}$. 
Fig.~\ref{fig:coordinates} illustrates this coordinate transformation.

Here, the approximate hole pose in the world coordinate, $^\mathrm{w}_\mathrm{h}T$, is available from the third assumption, and the end-effector pose with respect to the peg, $^\mathrm{p}_\mathrm{ee}T$, is known from the first assumption.
Therefore, the coordinate transformation is given by:
\begin{equation}\label{transformation}
    ^\mathrm{w}_\mathrm{ee}\hat{T} = \,^\mathrm{w}_{\mathrm{h}}T \, ^\mathrm{h}_\mathrm{p}\hat{T} \, ^\mathrm{p}_{\mathrm{ee}}T,
\end{equation}
where $^\mathrm{h}_\mathrm{p}\hat{T}$ denotes the inverse of $^\mathrm{p}_\mathrm{h}\hat{T}$, i.e., the pose of the peg with respect to the hole. Using the estimated $^\mathrm{w}_\mathrm{ee}\hat{T}$, the robot moves to the pre-insertion position, from which insertion is executed.

    \begin{figure*}[t]
    \centering
    \includegraphics[width=\textwidth]{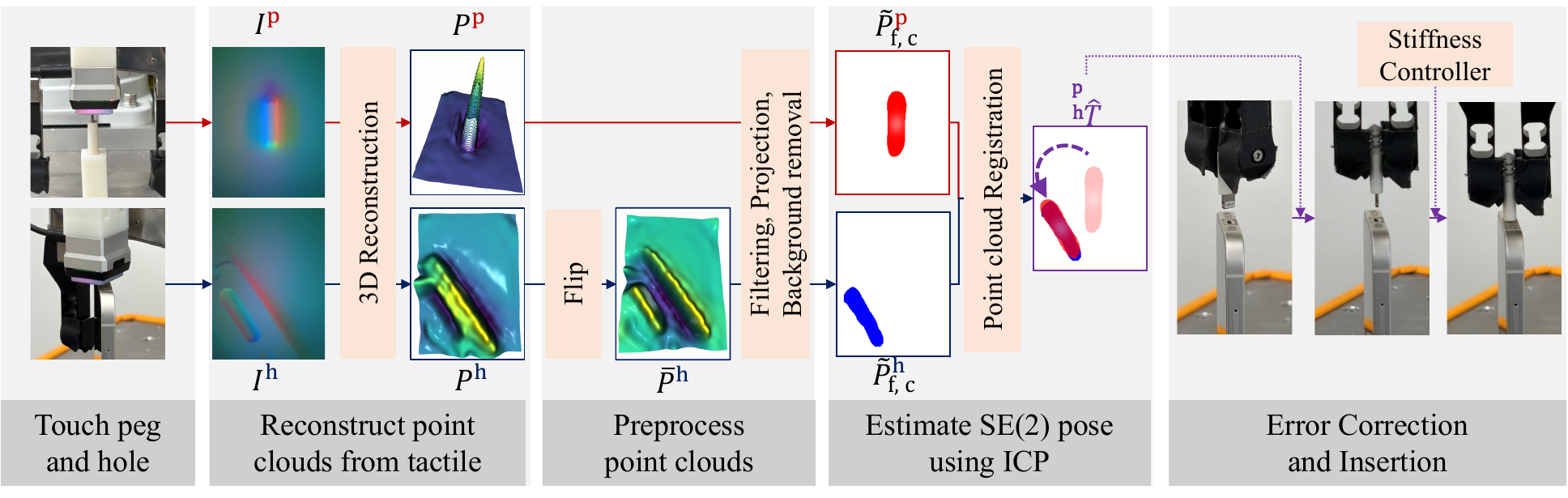}
    \caption{\textbf{Touch2Insert}: Overview of the proposed peg insertion framework. Tactile images are first converted into gradient maps and integrated to reconstruct 3D cross-sectional shapes of the peg and hole. The resulting point clouds are then refined by inverting the hole geometry, applying height-based filtering, projecting onto a 2D plane, and removing background artifacts. The cleaned planar point clouds are aligned using ICP with multiple initializations to estimate the relative $\mathrm{SE}(2)$ pose between peg and hole. Finally, the robot performs insertion under stiffness control, compensating for small residual errors without requiring exploratory search.
    }
    \label{fig:workflow}
\end{figure*}

\section{Method}\label{sec:system}
We propose \emph{Touch2Insert}, a complete system for connector insertion that estimates the $\mathrm{SE}(2)$ pose of a hole relative to a peg from vision-based tactile images and executes the subsequent insertion.
The key idea is to exploit tactile images as geometric observations, enabling accurate cross-sectional reconstruction and subsequent pose estimation without task-specific priors.
The overall pipeline consists of four stages:
(1) reconstructing 3D cross-sectional shapes of the peg and hole from tactile images,
(2) extracting regions of interest from the reconstructed point clouds,
(3) estimating the relative pose in $\mathrm{SE}(2)$ via ICP-based registration, and
(4) executing insertion using stiffness control for robustness. 
An overview of the Touch2Insert workflow is provided in Fig.~\ref{fig:workflow}, and the detailed procedure is summarized in Algorithm~\ref{alg:tac2insert}.

\subsection{Reconstruction of Cross-Sectional Shapes from Tactile Images} \label{subsec:3d_reconstruction}
A vision-based tactile sensor consists of a soft gel layer and an internal camera~\cite{yuan2017gelsight}. 
When the sensor makes contact with an object, the gel deforms and the camera records this deformation.
By processing the captured images, we can infer surface geometry at the contact region, which is essential for estimating peg–hole alignment.

To reconstruct the three-dimensional shape of the contact surface, we first estimate the gradient map $(\tfrac{\partial f}{\partial x}(x,y), \tfrac{\partial f}{\partial y}(x,y))$ from the tactile image $I(x,y)$, where $z=f(x,y)$ denotes the height map, and $x,y$ denote the position in pixel space. These gradient maps are then integrated by numerically solving a two-dimensional Poisson equation, yielding the height map $z=f(x,y)$~\cite{yuan2017gelsight}. The central challenge is how to accurately map tactile images to gradient maps. 

Existing approaches include photometric-stereo-based look-up tables~\cite{yuan2017gelsight} and multilayer perceptrons (MLPs)~\cite{wang2021gelsight} that map the pixel values to surface gradients. However, look-up tables discretize RGB values and leave residual errors, while MLPs ignore spatial correlations between pixels, limiting reconstruction accuracy.

To overcome these limitations, we employ a CNN-based model that predicts surface gradients from tactile images of the peg and hole, $I^\mathrm{p}$ and $I^\mathrm{h}$. By exploiting local spatial relationships, CNNs capture edges and fine structures more effectively, leading to more accurate shape reconstruction. We adopt a network design that incorporates the first and third layers of ResNet-50~\cite{he2016deep} to extract hierarchical features from tactile images.

The features are connected to a regression layer to estimate the gradients. Finally, the predicted gradient maps are converted into point cloud to obtain the 3D cross-sectional shapes of the peg and hole at the contact interface, denoted as $P^\mathrm{p}$ and $P^\mathrm{h}$.

For training, we use both real tactile images and simulated data to improve the model's generalization ability. For the real data, we obtain ground-truth gradient maps by pressing a sphere of known diameter against the sensor, annotating its center and radius, and computing the corresponding gradient maps~\cite{yuan2017gelsight,wang2021gelsight}. For the simulated data, we use Taxim~\cite{si2022taxim} to generate multiple simulated tactile images by pressing a cylindrical object that approximates the hole geometry against the sensor, in order to improve the accuracy and robustness of gradient estimation.

\subsection{Filtering of Point Cloud and Projection onto a Two-Dimensional Plane}\label{subsec:filter_pcd}
To enable reliable registration, the raw point clouds $P^\mathrm{p}$ and $P^\mathrm{h}$ reconstructed in the previous step are filtered and projected onto a two-dimensional plane.
This process removes background artifacts, resolves inconsistencies between peg and hole geometries, and yields clean cross-sectional shapes for pose estimation.
The procedure consists of four steps: 

\textbf{Inversion of the hole geometry.}
For consistent alignment, the hole point cloud is inverted along the $z$-axis, yielding $\bar{P}^\mathrm{h}$.
This transformation converts the concave geometry of the hole into a convex one, allowing the peg and hole point clouds to be compared in a consistent convex–convex form.

\textbf{Height-based filtering.}
As shown in Fig.~\ref{fig:workflow}, the raw point clouds include background regions that hinder accurate pose estimation.
To isolate the outer boundary, we apply height-based thresholding to the hole point cloud and remove its internal structures.
For both $P^\mathrm{p}$ and $\bar{P}^\mathrm{h}$, only the points below a threshold $z_{\mathrm{th}}$ are retained, producing the filtered point clouds $P_\mathrm{f}^\mathrm{p}$ and $\bar{P}_\mathrm{f}^\mathrm{h}$ that represent the convex shapes of the peg and hole, respectively.

\textbf{Projection onto a 2D plane.}
Since industrial connectors often contain internal concave/convex features that differ between the peg and hole, direct 3D registration can be unreliable. To address this, the height components of $P_\mathrm{f}^\mathrm{p}$ and $\bar{P}_\mathrm{f}^\mathrm{h}$ are set to zero, yielding planar point clouds $\tilde{P}_\mathrm{f}^\mathrm{p}$ and $\tilde{P}_\mathrm{f}^\mathrm{h}$ defined on the $(x,y)$ plane (with $z=0$).

\textbf{Background removal from the hole point cloud.}
Although $\tilde{P}_\mathrm{f}^\mathrm{p}$ now represents only the peg cross-section, $\tilde{P}_\mathrm{f}^\mathrm{h}$ may still include background artifacts, potentially leading to pose estimation failures.
To eliminate them, we apply DBSCAN clustering~\cite{schubert2017dbscan}.
The convex hull and area of each cluster are computed, and the largest-area cluster—corresponding to the background—is discarded.
This yields the refined hole point cloud $\tilde{P}_\mathrm{f, c}^\mathrm{h}$.

\subsection{Peg–Hole Registration}\label{subsec:registration}
To estimate the relative pose between the peg and hole, we perform two-dimensional ICP~\cite{besl1992icp} on the planar point clouds $\tilde{P}_\mathrm{f}^\mathrm{p}$ and $\tilde{P}_\mathrm{f, c}^\mathrm{h}$.
Formally, let
$$
\tilde{P}_\mathrm{f}^\mathrm{p} = \{\, p_\mathrm{p}^{(i)} \,\}_{i=1}^{N_\mathrm{p}}, 
\qquad 
\tilde{P}_\mathrm{f, c}^\mathrm{h} = \{\, p_\mathrm{h}^{(j)} \,\}_{j=1}^{N_\mathrm{h}},
$$
where $p_\mathrm{p}^{(i)}$ and $p_\mathrm{h}^{(j)}$ denote individual 2D points in the peg and hole clouds, respectively.
At iteration $t$, the correspondence is defined as
\begin{equation}
    \phi_t(j) = \arg\min_{i} \left\|\, p_\mathrm{h}^{(i)} - T_t p_\mathrm{p}^{(j)} \,\right\|,
\end{equation}
and ICP updates the transformation by solving
\begin{equation}
    \hat{T}_{t+1} = \arg\min_{T_{t} \in \mathbb{SE}(2)} 
\sum_{j=1}^{N_\mathrm{p}} 
\left\|\, p_\mathrm{h}^{\bigl(\phi_t(j)\bigr)} - T_t p_\mathrm{p}^{(j)} \,\right\|^2. 
\end{equation}

Since ICP is sensitive to initialization, we adopt a multi-initialization strategy.

Specifically, $\tilde{P}_\mathrm{f}^\mathrm{p}$ is rotated around its centroid by $\alpha \in \{ 0^\circ, 10^\circ, 20^\circ, \dots, 360^\circ \}$, and ICP is executed from each initialization.
Among the candidate results, the transformation with the largest number of inliers is selected.
Let $T^\ast$ denote the best transformation, $\theta^\ast$ its rotation component, and $\alpha^\ast$ the corresponding initial rotation.

The final rotation angle is then
\begin{equation}
    \hat{\theta} = \theta^{*} + \alpha^{*}.
\end{equation}
Accordingly, the transformation matrix $^\mathrm{p}_\mathrm{h}\hat{T}$ from $\tilde{P_\mathrm{f}^\mathrm{p}}$ to $\tilde{P}_\mathrm{f}^\mathrm{h}$ is obtained by replacing the rotation angle of $T^*$ with $\hat{\theta}$. That is, letting
\begin{equation}
    T^{*} = 
\begin{bmatrix}
R(\theta^{*}) & t^{*} \\
0 & 1 \\
\end{bmatrix},
\end{equation}
the transformation matrix $^\mathrm{p}_\mathrm{h}\hat{T}$ can be expressed as
\begin{equation}
    ^\mathrm{p}_\mathrm{h}\hat{T} = 
\begin{bmatrix}
R(\hat{\theta}) & t^{*} \\
0 & 1 \\
\end{bmatrix}.
\end{equation}
This transformation is subsequently used to move the peg to the pre-insertion position.

\begin{algorithm}[tb]
\caption{:\textit{Touch2Insert}}
\label{alg:tac2insert}
\hspace*{\algorithmicindent} \textbf{Input} Tactile image of peg $I^\mathrm{p}$, threshold $z_{\mathrm{th}}$ for filtering along the $z$-axis, maximum number of iteration of ICP $N^\mathrm{max}$

\hspace*{\algorithmicindent} \textbf{Output} Pre-insertion pose $^\mathrm{w}_\mathrm{ee}\hat{T}$

\begin{algorithmic}[1]
    \State Contact the cross-section of the hole and acquire a tactile image $I^{\mathrm{h}}$
    \State $P^{\mathrm{p}}, P^{\mathrm{h}} \gets \text{Reconstruct3D}(I^{\mathrm{p}}, I^{\mathrm{h}})$
    \State $\bar{P}^{\mathrm{h}} \gets \text{Flip}(P^{\mathrm{h}})$ along the z-axis
    \State $P^{\mathrm{p}}_{\mathrm{f}}, \bar{P}^{\mathrm{h}}_{\mathrm{f}} \gets \text{Remove points with } z \leq z_{\mathrm{th}}$
    \State $\tilde{P}^{\mathrm{p}}_{\mathrm{f}}, \tilde{P}^{\mathrm{h}}_{\mathrm{f}} \gets \text{Project to 2D by setting } z=0$ to all points
    \State $\tilde{P}^{\mathrm{h}}_{\mathrm{f,c}} \gets \text{Remove background}$
    \State $r_{\max} \gets 0$
    \For{$i \gets 1$ to $N^{\max}$}
        \For{$\alpha = 0$ \textbf{to} $360-\Delta\alpha$ \textbf{step} $\Delta\alpha$}
            \State $\tilde{P}^{\mathrm{p}}_{\mathrm{f},\alpha} \gets \text{Rotate } \tilde{P}^{\mathrm{p}}_{\mathrm{f}} \text{ by } \alpha \text{ degrees}$
            \State $\tilde{P}^{\mathrm{p}}_{\mathrm{f},\alpha,\mathrm{ICP}} \gets \text{ICP in } \mathrm{SE}(2)(\tilde{P}^{\mathrm{p}}_{\mathrm{f},\alpha}, \tilde{P}^{\mathrm{h}}_{\mathrm{f,c}})$
            \State $r_{\mathrm{in}} \gets \text{ComputeInlierRatio}(\tilde{P}^{\mathrm{p}}_{\mathrm{f},\alpha,\mathrm{ICP}}, \tilde{P}^{\mathrm{h}}_{\mathrm{f,c}})$
            \If{$r_{\mathrm{in}} > r_{\max}$}
                \State $r_{\max} \gets r_{\mathrm{in}}$
                \State $T^\ast \gets T_{\alpha}$
            \EndIf
        \EndFor
    \EndFor
    \State Compute ${}^\mathrm{w}_\mathrm{ee}\hat{T} = {}^\mathrm{w}_\mathrm{h}T T^{\ast} {}^{\mathrm{p}}_\mathrm{ee}T $
    \State \Return Pre-insertion pose of the end effector ${}^\mathrm{w}_\mathrm{ee}\hat{T}$
\end{algorithmic}
\end{algorithm}

\subsection{Insertion with Stiffness Controller}\label{subsec:stiffness_insertion}
Once the pre-insertion pose is obtained, the robot moves the peg to this position, located just above the hole, and the insertion process begins.
Some prior studies relying on vision alone achieved insertion by compensating for pose errors through exploratory motions such as spiral search; however, such strategies increase insertion time and reduce efficiency~\cite{yajima2025zeroshot}.
In contrast, Touch2Insert performs insertion directly with a stiffness controller~\cite{salisbury1980active}, which absorbs small residual errors and enables smooth alignment, thereby eliminating the need for additional exploratory searches.
Specifically, after reaching the pre-insertion position, the stiffness controller is activated and the peg is lowered to complete the insertion.

    \section{Experiments}\label{sec:results}
We conduct a series of experiments to evaluate the effectiveness of the proposed framework in connector insertion tasks. Our objectives are threefold: (1) to assess the accuracy of pose estimation in controlled simulation environments where ground-truth is available, (2) to validate the complete insertion pipeline in real-world settings with diverse connectors, and (3) to evaluate the accuracy of 3D reconstruction from tactile images. These experiments collectively examine whether our method can generalize to different connector geometries and operate reliably under realistic industrial conditions.
In these experiments, we used 45 real images and 69 simulated images for training. We also used a 4 mm metal sphere to obtain calibrated gradient maps.

\begin{figure}
    \centering
    \includegraphics[width=\linewidth]{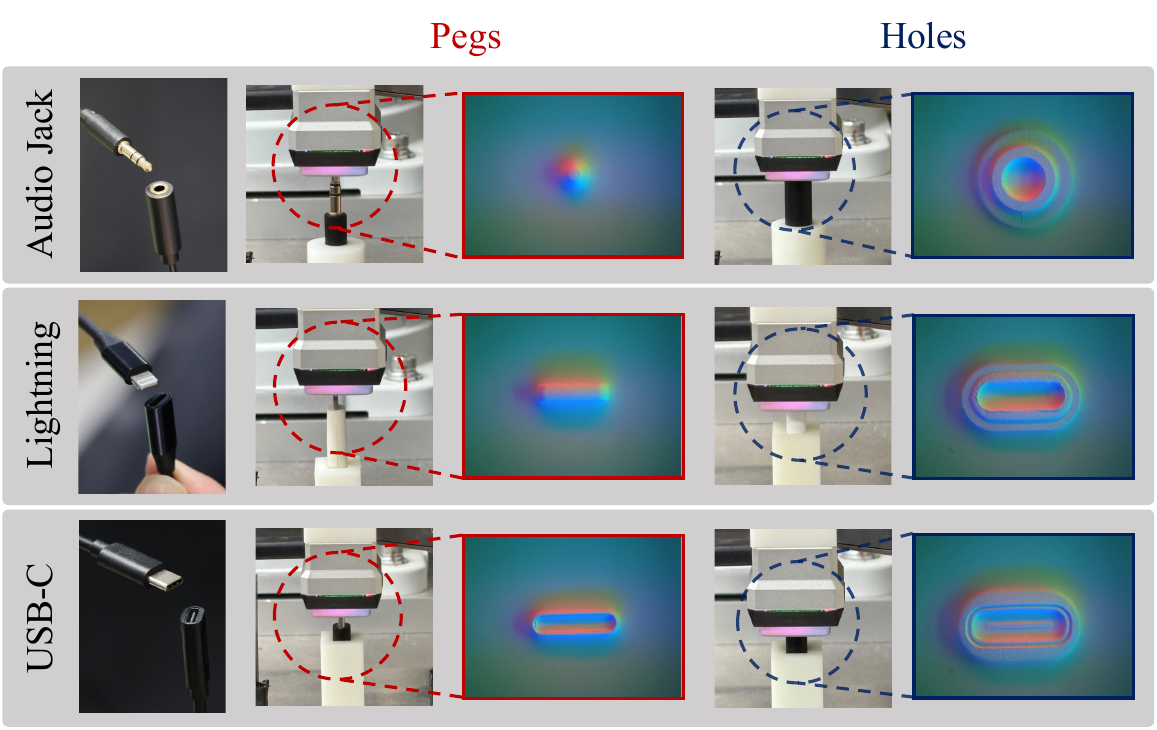}
    \caption{Connectors and example tactile images of the corresponding pegs and holes used in our experiments.}
    \label{fig:connectors}
\end{figure}

\subsection{Pose Estimation Performance in Simulation}\label{subsec:simulation_experiments}
We quantitatively evaluate the performance of the proposed pose estimation method. In real-world settings, obtaining ground-truth contact poses is difficult, as vision-based localization of the hole often suffers from a few millimeters of error due to jig misalignment or sensor calibration. Therefore, we conducted controlled evaluations in simulation using Taxim~\cite{si2022taxim}, a simulator for vision-based tactile sensors.

\textbf{Settings.}
We used CAD models of three types of connectors---Audio Jack, Lightning, and USB-C---as shown in ~\fref{fig:connectors}. Using Taxim~\cite{si2022taxim}, we simulated these models and generated virtual tactile images corresponding to the cross-sectional contact shapes. 
To mimic the localization errors that occur when the robot makes initial contact based only on vision, the hole poses were perturbed relative to the origin by $\Delta x,\Delta y \in [-4.0, 4.0]\,\mathrm{mm}$ in steps of $1.0\,\mathrm{mm}$, excluding $0.0\,\mathrm{mm}$, and $\Delta \theta \in [0^\circ, 315^\circ]$ in steps of $45^\circ$. This resulted in 512 pose variations in total.
Pose estimation of the holes with respect to the pegs was then performed.

For evaluation, symmetries of the connectors were taken into account: the rotation was ignored for the circular Audio Jack, and 180° rotational symmetry was considered for Lightning and USB-C.

\textbf{Baselines.}
To evaluate the effectiveness of our method, we compare it against two baselines. The first baseline, \emph{OmniGlue}, is based on image feature matching. Specifically, it  extracts edges from tactile images of the peg and hole using EDTER~\cite{pu2022edter}, performs feature matching with OmniGlue~\cite{jiang2024omniglue}, and estimates the relative pose from the resulting keypoint correspondences. 
We adopt this pipeline because directly estimating relative pose from tactile images is challenging.
The second baseline, \emph{w/o preprocess}, is introduced to evaluate the contribution of our preprocessing step, the filtering of point cloud and the projection onto a two-dimensional plane. In this baseline, only flipping is applied to the whole point cloud, while filtering and 2D projection are omitted. Registration is instead performed directly on the raw 3D point cloud, and the pose estimation accuracy in SE(2) is quantitatively evaluated.


\textbf{Metrics.}
Since successful connector insertion depends on both accurate positioning and orientation, we evaluate performance using the translation error $e_{\text{trans}}$ and the rotation error $e_{\text{rot}}$ between the estimated transformation and the ground-truth transformation.
Here, the translation components of the estimated and ground-truth transformation matrices are denoted by $\bm{t}_{\text{est}}$ and $\bm{t}_{\text{gt}}$, respectively, and the rotation components (in $\mathrm{SE}(2)$) by $\theta_{\text{est}}$ and $\theta_{\text{gt}}$.

The translation error is defined as the Euclidean distance between the two translation vectors in 2D:
\begin{equation}
    e_{\text{trans}} = \left\lVert \mathbf{t}_{\text{est}} - \mathbf{t}_{\text{gt}} \right\rVert_{2},
\end{equation}
and the rotation error is given by the absolute difference in angles:
\begin{equation}
    e_{\text{rot}} = |\theta_{\text{est}} - \theta_{\text{gt}}|.
\end{equation}

\begin{table}
    \centering
    \caption{Average estimation errors (mean values, with standard deviations in parentheses) for each connector. Bold numbers show the best results.}
    \begin{tabular}{rrcc} \toprule
         & & Translation  & Rotation \\
         & & Error (mm) & Error (degs) \\ \midrule
         \multirow{3}{*}{\emph{OmniGlue}}
         & \emph{Audio Jack} & 0.78 (0.39) & - \\
         & \emph{Lightning} & 2.25 (1.74)  & 31.64 (33.45) \\ 
         & \emph{USB-C} & 2.42 (2.13) & 28.29 (28.4)  \\ \midrule
         \multirow{3}{*}{\emph{w/o preprocess}}
         & \emph{Audio Jack} & 3.6 (1.21) & - \\
         & \emph{Lightning} & 3.68 (1.05)  & 44.56 (30.73) \\ 
         & \emph{USB-C} & 3.5 (1.20) & 45.47 (30.82)  \\ \midrule
         \multirow{3}{*}{Ours}
         & \emph{Audio Jack} & $\bm{0.56\ (0.23)}$ & - \\
         & \emph{Lightning} & $\bm{0.78\ (0.30)}$  & $\bm{4.36\ (4.44)}$ \\ 
         & \emph{USB-C} & $\bm{0.60\ (0.26)}$ & $\bm{2.43\ (2.14)}$  \\
         \bottomrule
    \end{tabular}
    \vspace{-3mm}
    \label{tab:result_estimation}
\end{table}

\begin{figure}[t]
    \centering
    \begin{subfigure}{0.45\linewidth}
        \includegraphics[width=\linewidth]{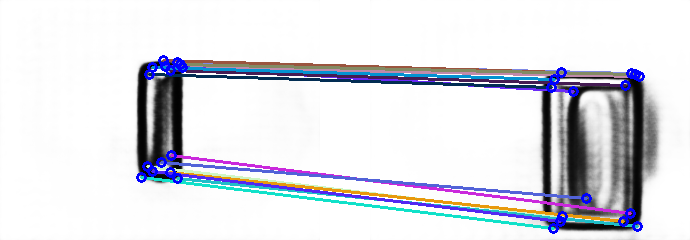}%
        \caption{Successful case}
    \end{subfigure}
    \begin{subfigure}{0.08\linewidth}
    \end{subfigure}
    \begin{subfigure}{0.45\linewidth}
        \includegraphics[width=\linewidth]{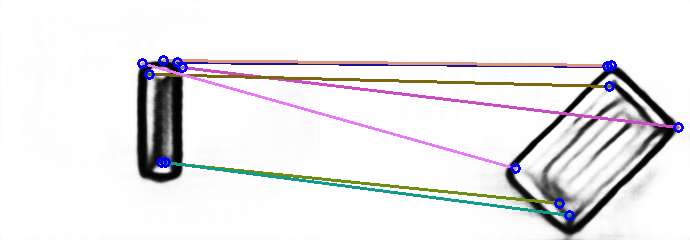}%
        \caption{Failed case}
    \end{subfigure}
    \caption{Successful and failed examples of \baselineomni\ baseline that uses edge detection and feature matching for the Lightning connector. For each image, the left shows the edge image of a peg, and the right shows a hole. In the successful case (a), matching occurs roughly at the corresponding positions of the peg and the hole. In contrast, in the failed case (b), there are regions where matching occurs at non-corresponding positions, leading to an inaccurate estimation.}
    \label{fig:Omniglue}
\end{figure}
\textbf{Results.}
Table~\ref{tab:result_estimation} shows the average estimation errors and standard deviations over 512 initial poses, demonstrating that the proposed method outperforms the baseline. Specifically, our method achieved an average translation error of less than $1\,\mathrm{mm}$ across all three types of connectors on average, and the rotation error was significantly smaller than that of the baselines.

Regarding the results of \emph{OmniGlue}, while it achieved sub-millimeter estimation accuracy for the audio jack, the error on the other two connectors was significantly larger than that of the proposed method in terms of both translation and rotation errors. \fref{fig:Omniglue} shows success and failure cases of pose estimation using \textit{OmniGlue} on the Lightning connector. In many of the \emph{OmniGlue} results, matching occurred at non-corresponding positions between the peg and the hole, which resulted in large estimation errors.

As for the \emph{w/o preprocess} setting, the accuracy was considerably lower than the proposed method for all connectors in both translation and rotation. This degradation is attributed to the ICP algorithm incorrectly registering overlapping background point clouds.

\begin{figure}[t]
    \centering
    \begin{subfigure}{0.4\linewidth}
        \includegraphics[width=\linewidth]{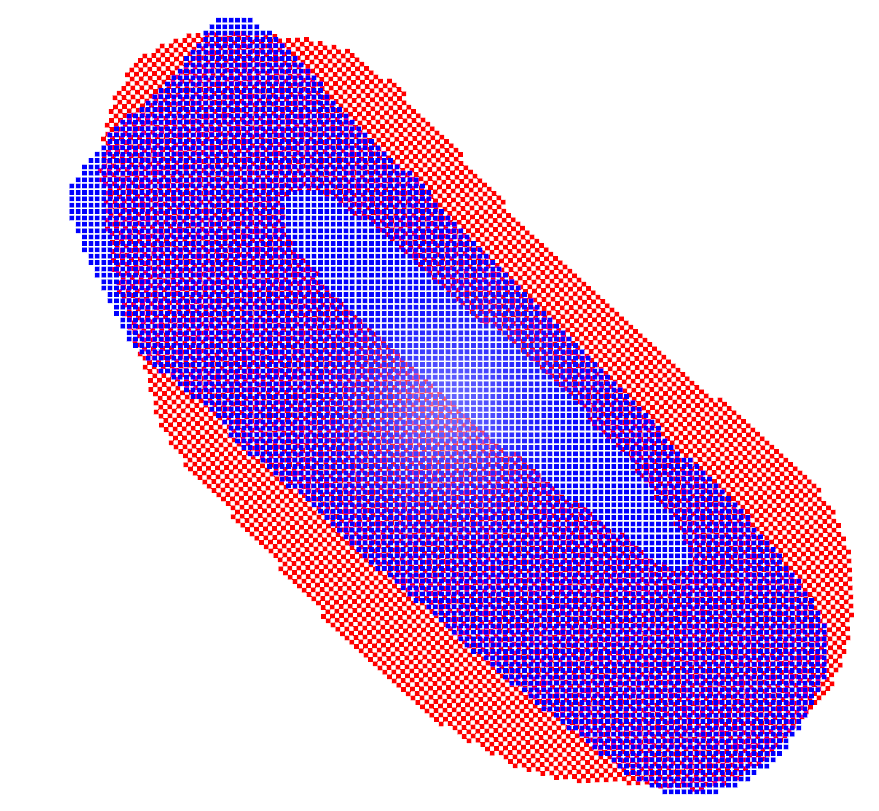}%
        \caption{Successful case}
    \end{subfigure}
    \begin{subfigure}{0.25\linewidth}
    \end{subfigure}
    \begin{subfigure}{0.25\linewidth}
        \includegraphics[width=\linewidth]{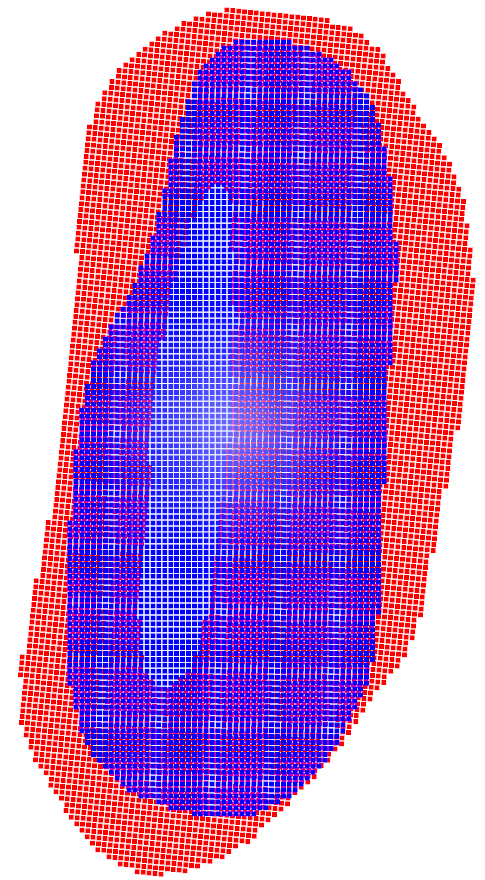}%
        \caption{Failed case}
    \end{subfigure}
    \caption{Registration results for successful and failed USB-C insertions. The red point cloud represents the peg and the blue point cloud represents the hole. In the failed case, distortions in the reconstruction produced irregularities in the hole point cloud, leading to a small angular misalignment with the peg and subsequent pose estimation errors.}
    \label{fig:failure_case}
\end{figure}

\subsection{Real-World End-to-End Evaluation with Insertion}
We next evaluate the complete insertion framework in a real environment. Unlike the simulation experiments in~\sref{subsec:simulation_experiments}, which tested pose estimation using idealized point clouds, this evaluation assesses the entire pipeline---from tactile image acquisition and 3D reconstruction to $\mathrm{SE}(2)$ pose estimation and physical insertion.

\textbf{Settings.}
Experiments were conducted using a MELFA RV-4FRL, a high-precision 6-DoF industrial robot equipped with a force/torque sensor for stiffness control. A Gelsight Mini tactile sensor~\cite{yuan2017gelsight} was mounted on the robot gripper via a 3D-printed support. The sensor provides $240 \times 320$ RGB images over an $18.6 \times 14.3$ mm sensing area, enabling the robot to capture cross-sectional images of the hole upon contact.
To emulate pose errors from vision-based hole localization, the hole pose was randomly perturbed before each trial:
$\Delta x, \Delta y \sim \mathcal{U}(-4.0, 4.0) [\mathrm{mm}]$ and $\Delta \theta \sim \mathcal{U}(0, 360)$ [$^\circ$].
The robot moves to the perturbed pose, acquires tactile images, estimates the hole pose relative to the peg in real time, and executes insertion using stiffness control. We evaluate three connector types (Fig.~\ref{fig:connectors}) with 20 trials each, for a total of 60 trials. This setup tests whether the proposed framework generalizes across diverse connector geometries under realistic conditions.

\textbf{Metrics.}
Performance was measured by insertion success rate, defined as the number of successful trials out of 20 per connector. This metric reflects the integrated performance of pose estimation and insertion.

\textbf{Results.}
Table~\ref{tab:result_insertion} reports the insertion success rates and completion times for each connector.
For the Audio Jack and Lightning connectors, the success rate exceeded 95\%, indicating that nearly all trials were successful.
Across all three connectors, the overall average reached 86.7\%, demonstrating that the proposed method is applicable to real-world insertion tasks.

\fref{fig:failure_case} shows registration results for a successful and failed insertion with a USB-C connector. The failure can be attributed to distortions in the 3D reconstruction, which caused a discrepancy in the size of the peg and hole point clouds, leading to slight misalignments in angle and position during point cloud registration.
Moreover, since USB-C has a smaller tolerance compared to the other two connectors, even a slight misalignment significantly affected the success rate. 

\begin{table}
    \centering
    \caption{Insertion success rate for each connector.}
    \begin{tabular}{rr} \toprule
         Connector Type & Success rate \\ \midrule
         \emph{Audio Jack} & 95\% (19/20)  \\
         \emph{Lightning} & 100\% (20/20) \\ 
         \emph{USB-C} & 65\% (13/20)  \\
         \bottomrule
    \end{tabular}
    \label{tab:result_insertion}
\end{table}

\subsection{Reconstruction Quality from Tactile Images}

\begin{table}[t]
    \centering
    \caption{Mean absolute errors (MAEs) between the predicted and ground-truth components $G_x$, $G_y$, $\theta_x$, and $\theta_y$ for gradient maps estimated by the baseline MLP and the proposed method. In each entry, the left and right values indicate the MLP-based model and the proposed model, respectively, and bold numbers show the better results. The proposed model achieves consistently lower errors, demonstrating more accurate tactile reconstruction.}
    
    \setlength{\tabcolsep}{2.5pt} 
    
    \resizebox{\columnwidth}{!}{
    \begin{tabular}{r c c c c c c c c} 
        \toprule
        \multirow{2}{*}{Connector Type} & 
        \multicolumn{2}{c}{\shortstack{$\mathrm{MAE}_{G_{x}}$ \\ \scriptsize{(mm/pixel)}}} & 
        \multicolumn{2}{c}{\shortstack{$\mathrm{MAE}_{G_{y}}$ \\ \scriptsize{(mm/pixel)}}} & 
        \multicolumn{2}{c}{\shortstack{$\mathrm{MAE}_{\theta_{x}}$ \\ \scriptsize{(degs)}}} & 
        \multicolumn{2}{c}{\shortstack{$\mathrm{MAE}_{\theta_{y}}$ \\ \scriptsize{(degs)}}} \\ 
        
        \cmidrule(lr){2-3} \cmidrule(lr){4-5} \cmidrule(lr){6-7} \cmidrule(lr){8-9}
        
        & MLP & Ours & MLP & Ours & MLP & Ours & MLP & Ours \\ 
        \midrule        
        
        \emph{Audio Jack Peg}  & 0.003 & $\bm{0.002}$ & 0.004 & $\bm{0.001}$ & 0.146 & $\bm{0.077}$ & 0.203 & $\bm{0.069}$ \\
        \emph{Audio Jack Hole} & 0.008 & $\bm{0.007}$ & 0.010 & $\bm{0.006}$ & 0.436 & $\bm{0.362}$ & 0.547 & $\bm{0.326}$ \\ 
        \emph{Lightning Peg}   & 0.005 & $\bm{0.004}$ & 0.007 & $\bm{0.003}$ & 0.274 & $\bm{0.196}$ & 0.369 & $\bm{0.177}$ \\
        \emph{Lightning Hole}  & 0.009 & $\bm{0.008}$ & 0.011 & $\bm{0.007}$ & 0.487 & $\bm{0.450}$ & 0.601 & $\bm{0.407}$ \\ 
        \emph{USB-C Peg}       & 0.008 & $\bm{0.007}$ & 0.010 & $\bm{0.007}$ & 0.457 & $\bm{0.400}$ & 0.560 & $\bm{0.374}$ \\
        \emph{USB-C Hole}      & $\bm{0.009}$ & 0.010 & 0.011 & $\bm{0.009}$ & $\bm{0.524}$ & 0.572 & 0.609 & $\bm{0.521}$ \\

        \bottomrule
    \end{tabular}
    }
    \label{tab:grad_estimation}
\end{table}

\begin{figure}[t]
    \centering
    \includegraphics[width=0.95\linewidth]{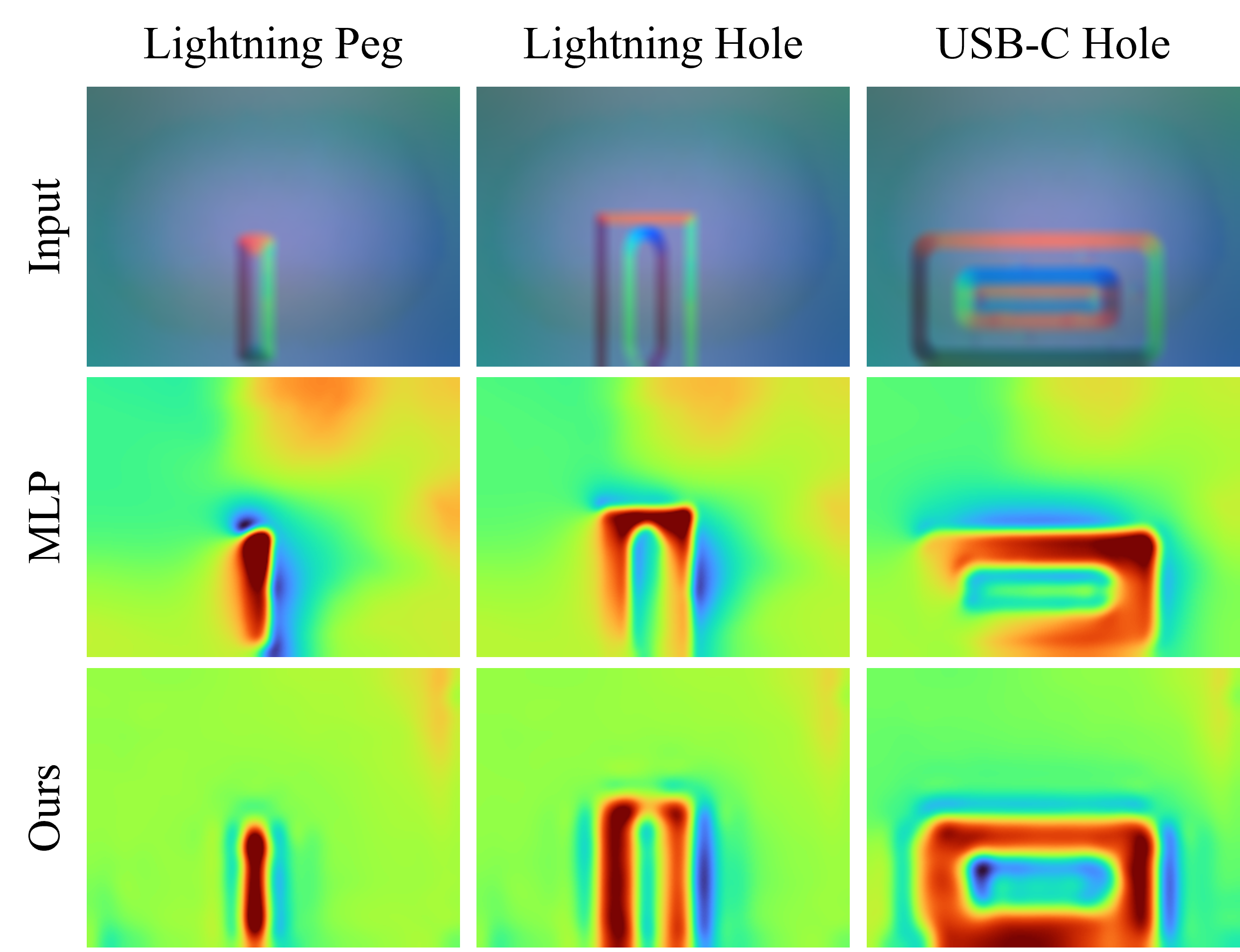}
    \caption{Height maps reconstructed from gradient maps estimated by the baseline MLP model and the proposed model, shown at the same scale. The proposed method produces cleaner and more accurate height maps than the MLP baseline, highlighting the effectiveness of our reconstruction approach and its contribution to reliable connector insertion.
    }
    \label{fig:height_result}
\end{figure}

\textbf{Settings.}
In this experiment, we eavluated the quality of the reconstructed gradient map against an MLP-based model~\cite{wang2021gelsight}. We generated pairs of tactile images and their corresponding ground truth gradient maps in simulation using Taxim~\cite{si2022taxim}. Specifically, as in~\sref{subsec:simulation_experiments}, we created data by making contact from 512 different initial positions. We then input the simulated tactile images into both the proposed model and an MLP model~\cite{wang2021gelsight} to estimate gradient maps, and evaluated the performance of the two models. In the evaluation settings, we normalized the norms to the range [0,1] in order to align the scale of each model’s outputs with the ground truth before computing the error.

Following Wang et al.~\cite{wang2021gelsight}, the input to the MLP is a difference image obtained by subtracting a blank background image from the raw image, as this provides a more stable and accurate mapping. In contrast, our ResNet-based method uses the captured images directly for better estimation accuracy.

\textbf{Metrics.}
We evaluate the mean absolute error (MAE) of the surface gradients
$G_k = \partial z / \partial k$ and the corresponding slope angles
$\theta_k = \arctan(G_k)$, where $k \in \{x, y\}$:
\begin{equation}
\mathrm{MAE}_{\phi} = \frac{1}{N} \sum_{i=1}^{N}
\left|\phi_{\mathrm{pred}}^{(i)} - \phi_{\mathrm{gt}}^{(i)}\right|,
\quad \phi \in \{G_k, \theta_k\}.
\end{equation}
Here, $N = HW$ is the total number of pixels for an image of size $H \times W$.

\textbf{Results.}
Table~\ref{tab:grad_estimation} compares the reconstruction errors of the MLP baseline and the proposed method for the peg and hole across three connector types. The left and right columns report the results of the MLP-based model and the proposed model, respectively.

For most connector types, the proposed method consistently achieves lower errors, indicating improved reconstruction accuracy. This improvement is consistent with the observation that the MLP baseline exhibits more widespread background distortions in the estimated gradient maps, whereas the proposed method yields a flatter and less distorted background (see \fref{fig:height_result}). This is likely because the CNN can leverage broader spatial context, which helps reduce estimation errors in background regions with gradual intensity changes.

By contrast, in contact regions where intensity changes are more pronounced, the two models can show similar errors in terms of the averaged metrics. However, as shown in \fref{fig:height_result}, the MLP model occasionally exhibits severe local reconstruction failure in parts of the contact region, resulting in missing contact geometry. Although such local defects may have only a limited effect on the averaged error, they can significantly affect the subsequent post-processing and may ultimately lead to insertion failures.    
\section{Conclusion and Future Work}\label{sec:conclusion}
In this study, we have presented \emph{Touch2Insert}, a tactile-based framework for arbitrary peg insertion that reconstructs cross-sectional geometry from tactile images and estimates the relative pose between peg and hole in $\mathrm{SE}(2)$. By aligning the reconstructed shapes through ICP, our method enables insertion from a single contact without task-specific training. Experiments in both simulation and on a real robot demonstrated that the framework achieves sub-millimeter pose estimation accuracy and an average success rate of 86.7\% across multiple connector types, confirming its effectiveness under the stringent tolerances required in industrial settings.

Looking ahead, we plan to extend the framework in several directions. First, we aim to relax the current assumption that the entire hole shape can be captured from a single contact. For holes larger than the sensor's field of view, we will investigate strategies that combine multiple contacts to reconstruct the complete geometry and still enable reliable insertion. Second, we intend to generalize the system into a multimodal framework by integrating tactile sensing with vision and force feedback, thereby increasing robustness and practicality in diverse scenarios. Finally, we seek to remove the reliance on a predefined grasping pose, which in this study was ensured by a jig. Our goal is to enable the robot to autonomously estimate the peg's grasping pose and execute insertion from arbitrary initial conditions, moving toward an end-to-end connector insertion pipeline that operates flexibly in real-world environments.

    \bibliographystyle{IEEEtran}
    \bibliography{references}

\end{document}